# From Model Choice to Model Belief: Establishing a New Measure for LLM-Based Research

Hongshen Sun[*]    Juanjuan Zhang[†]

August 20, 2025

**Abstract**

Large language models (LLMs) are increasingly used to simulate human behavior, but common practices to use LLM-generated data are inefficient. Treating an LLM's output ("model choice") as a single data point underutilizes the information inherent to the probabilistic nature of LLMs. This paper introduces and formalizes "model belief," a measure derived from an LLM's token-level probabilities that captures the model's belief distribution over choice alternatives in a single generation run. The authors prove that model belief is asymptotically equivalent to the mean of model choices (a non-trivial property) but forms a more statistically efficient estimator, with lower variance and a faster convergence rate. Analogous properties are shown to hold for smooth functions of model belief and model choice often used in downstream applications. The authors demonstrate the performance of model belief through a demand estimation study, where an LLM simulates consumer responses to different prices. In practical settings with limited numbers of runs, model belief explains and predicts ground-truth model choice better than model choice itself, and reduces the computation needed to reach sufficiently accurate estimates by roughly a factor of 20. The findings support using model belief as the default measure to extract more information from LLM-generated data.

---

[*]Ph.D. Student in Marketing, MIT Sloan School of Management, email: sunhs@mit.edu.
[†]John D. C. Little Professor of Marketing, MIT Sloan School of Management, email: jjzhang@mit.edu.

"Do I contradict myself? Very well then I contradict myself, (I am large, I contain multitudes.)" Walt Whitman's *Song of Myself* passage illustrates the challenge of understanding human thought: every human choice is just one draw from a complex distribution of often contradictory states of mind. For decades, substantial research efforts in marketing and other social sciences have sought to go beyond observed choices to infer the mental landscape from which they arise (e.g., Guadagni and Little 1983; McFadden 1986).

The advent of large language models (LLMs) offers a vast new data source for these human-centric research fields. LLMs are increasingly used as "silicon agents" to simulate a wide spectrum of human behaviors at scale (e.g., Argyle et al. 2023). When properly conditioned, LLMs can reveal foundational patterns of human behavior, such as price sensitivity in product purchase decisions, producing synthetic data with the potential to mirror real-world observations (e.g., Arora, Chakraborty, and Nishimura 2025; Brand, Israeli, and Ngwe 2024; Horton 2023).

In this paper, we highlight a unique advantage of LLM-generated data: their ability to go beyond the model's choice to offer a glimpse into the model's "mind"—a modern response to the age-old challenge of studying human thought. This advantage is rooted in the probabilistic nature of LLMs (Bengio et al. 2003). When prompted to make a choice, an LLM computes a probability distribution over potential responses internally, and then samples from this distribution to generate an output. This distributional knowledge over the potential choices offers deeper insights into the LLM's inner trade-offs than its eventual choice alone.

For example, suppose an LLM is prompted to make a choice between two diaper brands, Pampers and Huggies. The LLM computes the choice probabilities between the two brands. For simplicity of illustration, assume temporarily that the LLM selects the next token using greedy search, outputting the choice that is more probable (Minaee et al. 2024). While the LLM's eventual choice is binary, the underlying choice probabilities are more granular—they can be interpreted as the LLM's "mind shares" between the two brands, potentially derived from aggregated market shares in the consumer data used for training. To see the value of this more granular data, imagine two scenarios following a price drop by Pampers:

1. The LLM changes its probability of choosing Pampers from .49 to .51, and changes its eventual choice from Huggies to Pampers.

2. The LLM changes its probability of choosing Pampers from .01 to .49, and maintains its eventual choice of Huggies.

Using the LLM's eventual choice alone would likely overestimate the LLM's price sensitivity (in absolute value) in the first scenario and underestimate its price sensitivity in the second. By contrast, using the probability distribution provided by LLMs helps preserve the information that would otherwise be lost in the coarse aggregation process often involved in translating mind to choice.

This advantage of LLM-generated data has not been fully utilized in current practice. Common applications have often defaulted to treating LLMs like human subjects, recording a single response as the answer used for subsequent analysis. A subset of studies use resampling, repeatedly querying the model to derive a distribution of responses (e.g., Argyle et al. 2023; Brand, Israeli, and Ngwe 2024). The resampling approach is intuitively appealing, but can be computationally costly. The methodological gap, therefore, is the lack of a formal measure that captures the LLM's internal state efficiently.

The paper bridges this gap by introducing *model belief*, a measure derived from an LLM's token-level log-probabilities (logits). We formalize this measure to delineate the latent choice distribution over a set of alternatives, as opposed to *model choice*, the commonly used measure that records the model's eventual choice. We first prove a series of desirable theoretical properties of model belief. We show that it is asymptotically equivalent to the mean of unbiased samples of model choice (a non-trivial property, as we shall explain), but forms a more efficient estimator with lower variance and faster convergence. Analogous properties hold for quantities that are smooth functions of model belief and model choice—a useful result for downstream applications. To examine the empirical performance of model belief, we conduct a demand estimation study where an LLM simulates consumer responses to different prices. In limited numbers of runs, model belief explains and predicts ground-truth model choice better than model choice itself, with more accurate, precise, and robust price sensitivity estimates. To achieve sufficiently accurate estimates, model belief needs only about 1/20th of the computation required by model choice. Last, we discuss why model belief should be used as the default measure to extract more information from LLM-based research while doing so at lower costs.

The relationship between model belief and model choice echoes the relationship between latent utility, or preference intensity, and observed choice (Hauser and Shugan 1980). We adopt the term "model belief" to clarify its reference to the model's internal estimate of the probabilities across alternatives—that is, how likely the model believes each possible value of the next token should be. More substantively, while preference-intensity data can be more informative than observed choice data, they can also be more difficult to elicit from human respondents in market research (Hauser and Shugan 1980). In contrast, LLMs compute model beliefs first and then draw model choices accordingly. As such, model belief data are worth collecting for their additional informational value at minimal additional cost.

This paper is situated within the fast-growing literature across the social sciences using LLMs to simulate human responses (e.g., Aher, Arriaga, and Kalai 2023; Argyle et al. 2023; Demszky et al. 2023; Dillion et al. 2023; Horton 2023). Notably, the marketing literature has adopted LLMs to enhance marketing research (Arora, Chakraborty, and Nishimura 2025; Blanchard et al. 2025), quantify consumer preferences (Brand, Israeli, and Ngwe 2024), generate creative ideas (De Freitas, Nave, and Puntoni 2025), study intertemporal choices (Goli and Singh 2024), enable causal inference (Gui and Toubia 2025), conduct brand-perceptual analysis (Li et al. 2024), extract customer needs (Timoshenko, Mao, and Hauser 2025), and identify engaging content (Ye, Yoganarasimhan, and Zheng 2025). A focus of this literature is to examine the external validity of LLMs compared with humans, besides identifying useful applications. We contribute to this literature along a different dimension, focusing instead on internal measurement validity and information retrieval. The model belief measure we emphasize has the potential to enhance a wide range of LLM applications, across different external validity contexts, by enabling more efficient extraction of information from each LLM-based study.

# MODEL BELIEF: THEORETICAL FOUNDATIONS

This section lays the formal groundwork of our proposed measure for LLM-based studies. We begin by defining two key measures derived from LLM outputs: model choice, the standard single outcome from a generation run, and model belief, our proposed measure based on the LLM's token probabilities. We then prove two central properties: first, model belief and the mean of model choice are asymptotically equivalent, and second, model belief forms a statistically more efficient estimator. We also prove that analogous properties hold for smooth functions of model belief and model choice that are often used in downstream applications.

## *From Model Choice to Model Belief: Formal Definitions*

Let the input prompt for a generation run of an LLM be the token sequence $\mathbf{x}_{1:T_0} = (x_1, \cdots, x_{T_0})$, where each token belongs to the fixed vocabulary $\mathcal{V}$. (In the remainder of the mathematical presentation, boldface denotes vectors.) The LLM extends the input prompt sequence one token at a step in an auto-regressive loop. At each step $t = 1, 2, \ldots$, the model emits a vector of logits $\mathbf{z}_t \in \mathbb{R}^{|\mathcal{V}|}$ over $\mathcal{V}$. It is standard practice to then use the softmax operator to transform these logit scores into a vector of probabilities $\mathbf{P}_t = \mathrm{softmax}(\mathbf{z}_t) \in [0, 1]^{|\mathcal{V}|}$ over $\mathcal{V}$ (Vaswani et al. 2017), where

$$\mathbf{P}_t := (P_t(v))_{v \in \mathcal{V}}, \quad P_t(v) = \frac{\exp z_t(v)}{\sum_{v' \in \mathcal{V}} \exp z_t(v')}, \quad \forall v \in \mathcal{V}.$$

The next token $y_t$ is drawn from the distribution $\mathbf{P}_t$ following a sampling rule (Minaee et al. 2024). For example, greedy search outputs the most probable token. Temperature scaling adjusts the probabilities, where a temperature of 0 replicates greedy search, a temperature of 1 samples the next token in proportion to $\mathbf{P}_t$, and a higher temperature draws more evenly over the vocabulary. The chosen token is appended to the context, and the cycle repeats until the model produces an end-of-sequence symbol or a length cap is reached.

In decision-focused studies, we often do not need the entire output sequence. Instead, the focus is often on the "pivot token:" the first generated token that unambiguously resolves the model's decision among a predefined set of mutually exclusive alternatives, $\mathcal{J} \subseteq \mathcal{V}$. Denote the pivot token as $y_\tau \in \mathcal{J}$, where $\tau$ denotes its corresponding position in the output sequence. We formally define model choice as

$$c := y_\tau \in \mathcal{J}$$

to denote the concrete selection made by the model on a generation run given the input.[1]

While the model choice $c$ represents the final decision outcome from a single generation run, our premise is that it is often valuable to access a more nuanced measure of the model's underlying belief distribution across the alternatives before this single choice is actualized. This leads to the concept of model belief. Because the model decision $c$ is sampled directly from $\mathbf{P}_\tau$, the model's ex-ante belief over the choice alternatives in $\mathcal{J}$ is completely captured by the sub-distribution

$$\boldsymbol{b} := (b(j))_{j \in \mathcal{J}}, \quad b(j) = \frac{P_\tau(j)}{\sum_{j' \in \mathcal{J}} P_\tau(j')}, \quad \forall j \in \mathcal{J},$$

where $P_\tau(j)$ is the probability of alternative $j$ at the pivot position $\tau$. We define this distribution $\boldsymbol{b}$ as model belief. The vector $\boldsymbol{b}$ measures the model's probabilistic belief of what the choice should be among the set of alternatives $\mathcal{J}$ at the precise moment the choice $c$ is made. Model belief offers an intuitive interpretation in application fields such as marketing. For example, in a brand choice context, $b(j)$ could be interpreted as the LLM's perceived market share of brand $j$ in the choice set $\mathcal{J}$, whereas $c$ only indicates which brand is chosen for a single generation run. This simple example suggests a potential advantage of using model belief as a measure with richer information than model choice. The next section will formally compare these two measures.

*Asymptotic Equivalence and Statistical Efficiency*

Building on the intuition that model belief may offer richer insights than model choice, we compare their statistical performance over multiple independent generation runs given a fixed input prompt. We take this approach for three reasons. First, the intuitive appeal of model belief over model choice, as illustrated thus far, pertains to a single generation run. A natural question is whether model choice would perform just as well when aggregated over many generation runs. Second, this approach captures current research practice. To reliably uncover a model's decision-making patterns, some researchers have used resampling, performing repeated runs to generate a distribution of model choices (e.g., Brand, Israeli, and Ngwe 2024). Third, this approach allows us to construct robust measures of both model choice and model belief by aggregating over the idiosyncrasies of individual generation runs and then evaluate their properties as estimators.

---

[1] This framework accommodates multi-token alternatives, where we generalize the definition of the pivot token to be the shortest sequence of initial tokens that uniquely maps to an alternative in $\mathcal{J}$, thereby unambiguously resolving the model's decision.

The comparison between the model choice and model belief estimators is not immediately clear. Even though the input prompt remains fixed, each generation run is a sequential sampling process itself. Consequently, the pivot position $\tau$ (and the sequence of tokens preceding the pivot token) can vary from run to run. For instance, queried with "*Would you choose to buy Pampers, Huggies, or neither?*", one run might produce response "*I would choose to buy **Pampers** because their overnight line...*" while another run may yield "*Based on the information provided, I would choose **Huggies**.*" Therefore, the comparison between model choice and model belief is more nuanced than simply reproducing a probability distribution through repeated independent draws. We present the formal comparison below.

Let $r = 1, \ldots, N$ index an LLM's generation runs (or runs, for brevity, hereafter). Write $c^{(r)}$ and $\boldsymbol{b}^{(r)}$ for the model choice and model belief on run $r$, respectively. Meanwhile, define the vector of true choice distribution across the choice set $\mathcal{J}$ as

$$\bar{\boldsymbol{c}} := (\bar{c}(j))_{j \in \mathcal{J}}, \quad \bar{c}(j) := \Pr(c = j),$$

where Pr denotes probability. Intuitively, $\bar{c}(j)$ tells us how likely the LLM will choose alternative $j$ when the same input prompt is repeated infinitely. In practice, we approximate this choice distribution with its empirical analogue, where

$$\hat{c}^{(N)}(j) = \frac{1}{N} \sum_{r=1}^{N} \mathbb{1}\{c^{(r)} = j\}.$$

As $N$ grows, the law of large numbers guarantees that $\hat{c}^{(N)}(j)$ converges almost surely to the true probability $\bar{c}(j)$.

Model belief offers a finer-grained view. Instead of waiting to see which token is ultimately sampled in a run, it provides the probabilities of all the alternatives at the pivot step $\tau$ in each run. As explained, the pivot step $\tau$ may vary across runs and should strictly be $\tau^{(r)}$ for run $r$; we drop the run superscript to facilitate exposition in the rest of this section. For run $r$, write the probabilities at the pivot step $\tau$ as

$$P_\tau^{(r)}(j) = \Pr\left(y_\tau^{(r)} = j \mid y_1^{(r)}, \cdots, y_{\tau-1}^{(r)}\right), \quad \forall j \in \mathcal{J}.$$

Importantly, the probabilities at the pivot step of a run depend on the sequence of tokens generated before the pivot step of that run (besides the input prompt), which in turn depend on the sampling strategy and, for nondeterministic sampling, the realization of the sampling draw.

Based on the probabilities specified above, model belief for run $r$ is defined as $\boldsymbol{b}^{(r)} := (b^{(r)}(j))_{j \in \mathcal{J}}$, where

$$b^{(r)}(j) = \frac{P_\tau^{(r)}(j)}{\sum_{j' \in \mathcal{J}} P_\tau^{(r)}(j')}, \quad \forall j \in \mathcal{J}.$$

True model belief is then defined as the expectation of the single-run model belief across all runs, so that

$$\bar{b}(j) := \mathbb{E}\left[b^{(r)}(j)\right]$$

with the corresponding empirical estimator being

$$\hat{b}^{(N)}(j) = \frac{1}{N} \sum_{r=1}^{N} b^{(r)}(j).$$

As we illustrated with the LLM diaper choice example, because model belief is derived from the model's internal probabilities at a specific and potentially variable pivot position $\tau$ for each run, its asymptotic equivalence to the empirically observed model choice distribution is not self-evident. Next, we provide a proposition and proof of this asymptotic equivalence result.

**Proposition 1** (Asymptotic Equivalence Between Model Belief and Model Choice). *Under unbiased sampling strategies,*[2]

$$\bar{b}(j) = \bar{c}(j), \quad \forall j \in \mathcal{J}.$$

*Proof.* Let $\mathcal{G}^{(r)} := (y_1^{(r)}, \cdots, y_{\tau-1}^{(r)})$ denote the sequence of tokens sampled up to run $r$'s pivot step $\tau$. This run's model choice is $c^{(r)} = y_\tau^{(r)}$. By definition, $\tau$ is the first step at which the model's decision is declared and the generated token $y_\tau^{(r)}$ must belong to the predefined set of alternatives $\mathcal{J}$. Thus, conditional on $\mathcal{G}^{(r)}$, the probability of choosing $j$ for run $r$ equals the probability of $y_\tau^{(r)} = j$ normalized by the total probability of selecting any alternative from $\mathcal{J}$ at step $\tau$:

$$\Pr\left(c^{(r)} = j \mid \mathcal{G}^{(r)}\right) = \frac{P_\tau^{(r)}(j)}{\sum_{j' \in \mathcal{J}} P_\tau^{(r)}(j')} = b^{(r)}(j).$$

Taking expectations on both sides and applying the law of total expectation, we have:

$$\bar{b}(j) = \mathbb{E}\left[b^{(r)}(j)\right] = \mathbb{E}\left[\Pr\left(c^{(r)} = j \mid \mathcal{G}^{(r)}\right)\right] = \Pr(c = j) = \bar{c}(j).$$

□

This asymptotic equivalence result is a reassuring property. It implies that unbiased resampling of model choices, when repeated sufficiently many times, can approximate the model's underlying probability distribution over the choice alternatives. The question, then, is whether the same goal can be achieved more efficiently using the model belief estimator. The answer is yes. The model belief estimator exhibits a smaller variance and a faster convergence rate than that of model choice. The following proposition formalizes this result.

**Proposition 2** (Greater Efficiency of the Model Belief Estimator). *Let $\hat{b}^{(N)}(j)$ and $\hat{c}^{(N)}(j)$ be the empirical model-belief share and model-choice share, respectively, of alternative $j \in \mathcal{J}$ after $N$ independent runs. Under unbiased sampling, the following properties hold.*

---

[2] Unbiased sampling means sampling tokens in proportion to their underlying probability distribution. In practice, unbiased sampling can be achieved by setting the temperature to 1 and using the full sampling dictionary.

1. **Smaller Sampling Variance.**

$$\text{Var}[\hat{b}^{(N)}(j)] \leq \text{Var}[\hat{c}^{(N)}(j)], \quad \forall j \in \mathcal{J}$$

   *with equality if and only if model belief is almost surely deterministic for alternative $j$.*

2. **Faster Convergence.** *For any target accuracy tolerance $\epsilon > 0$ and confidence level $1 - \delta$ with $0 < \delta < 1$, let $N_b$ be the minimal number of runs to guarantee $\Pr\left(|\hat{b}^{(N)}(j) - \bar{b}(j)| \leq \epsilon\right) \geq 1 - \delta$, and let $N_c$ be the minimal number of runs to guarantee $\Pr\left(|\hat{c}^{(N)}(j) - \bar{c}(j)| \leq \epsilon\right) \geq 1 - \delta$. The following holds:*

$$N_b \leq N_c$$

   *with equality if and only if model belief is almost surely deterministic for alternative $j$.*

*Proof.* See the Appendix.

In LLM applications, researchers often care about downstream quantities such as market shares, price sensitivity, and utility weights for various product features. The calculation of these quantities may involve a smooth function of model choice. For example, price sensitivity can be calculated based on how model choice changes in response to a price change in the input prompt. A useful result for LLM applications, as we will show, is that such downstream quantities derived from model belief are not only asymptotically equivalent to those derived from the model choice distribution, but also more statistically efficient.

Formally, let $g : \Delta^{|\mathcal{J}|} \to \mathbb{R}^q$ be any continuously differentiable mapping, where $\Delta$ is the probability simplex and $|\mathcal{J}|$ is the number of choice alternatives. Let $\theta_b = g(b)$ denote the $q$-dimensional downstream vector of interest when computed from model belief and $\theta_c = g(c)$ denote the counterpart when computed from model choice. Further, let $\bar{\theta} = g(\bar{c})$ denote the true downstream vector value computed from the true model choice distribution. Last, based on $N$ independent runs, define the plug-in estimators

$$\hat{\theta}_b^{(N)} := g\left(\hat{b}^{(N)}\right) \quad \text{and} \quad \hat{\theta}_c^{(N)} := g\left(\hat{c}^{(N)}\right),$$

where $\hat{\boldsymbol{b}}^{(N)} := (\hat{b}_j^{(N)})_{j \in \mathcal{J}}$ and $\hat{\boldsymbol{c}}^{(N)} := (\hat{c}_j^{(N)})_{j \in \mathcal{J}}$. These plug-in estimators satisfy the following properties, parallel to those in Propositions 1 and 2.

**Proposition 3** (Consistency of Plug-in Estimators). *With $N$ independent runs and unbiased sampling, the plug-in estimators based on model belief and model choice converge in probability to the true parameter values:*

$$\hat{\theta}_b^{(N)} \xrightarrow{p} \bar{\theta} \quad \text{and} \quad \hat{\theta}_c^{(N)} \xrightarrow{p} \bar{\theta}.$$

*Proof.* By the law of large numbers, we have $\hat{\boldsymbol{b}}^{(N)} \xrightarrow{p} \bar{\boldsymbol{b}}$ and $\hat{\boldsymbol{c}}^{(N)} \xrightarrow{p} \bar{\boldsymbol{c}}$. By Proposition 1, we have $\bar{\boldsymbol{b}} = \bar{\boldsymbol{c}}$. The consistency result is then proved by the continuous mapping theorem. □

**Proposition 4** (Greater Efficiency of Plug-in Estimators Based on Model Belief). *Under unbiased sampling, the plug-in estimator based on model belief has weakly smaller sampling covariance (in terms of Loewner dominance) and weakly faster convergence (in Euclidean space) than the plug-in estimator based on model choice, with equality if and only if model belief is almost surely deterministic for at least one alternative.*

*Proof.* See the Appendix.

In summary, we have shown that the model belief measure is theoretically superior to model choice. With an infinite number of LLM runs, model belief and model choice converge in probability to the same true value. However, the model belief estimator is statistically more efficient, with a smaller sampling variance and faster convergence speed. Analogous properties hold for downstream quantities that are smooth functions of model belief and model choice, respectively. We examine the empirical advantages of model belief in the following section.

## EMPIRICAL STUDY: LLM-BASED DEMAND ESTIMATION

In this section, we present an empirical comparison of model belief and model choice. Although their asymptotic equivalence and the greater efficiency of model belief are theoretically established properties, this empirical study nevertheless adds value for the following reasons. First, it helps visualize the magnitude of model belief's efficiency gains in a concrete application setting. Second, it helps examine the claim that, in finite samples typical of applied research, model belief is not only a more efficient but also a more accurate and robust measure due to its finer granularity. We illustrated the intuition behind this claim with the opening example on price-sensitivity estimation, where model belief goes beyond the discrete outcome of model choice to offer a continuous measure of preference strength. We now empirically demonstrate how this intuition leads to model belief's accuracy gain in a well-defined application context.

We focus on the context of demand estimation within a discrete-choice framework. This is a canonical and arguably one of the most important problems in marketing research. Demand estimation quantifies how consumer decisions respond to changes in prices and other product characteristics, which are essential inputs for answering a wide range of positive and normative questions in marketing and related fields ([Chintagunta and Nair 2011](#)). Discrete choice modeling is a foundational and widely used approach in demand estimation ([Guadagni and Little 1983](#); [McFadden 1986](#)). It captures the ubiquitous setting in which consumers choose from a set of alternatives based on their (often continuous) latent utilities, a real-world analogy to how model belief translates into model choice.

To implement this study, we present an LLM with different choice scenarios that vary in product price, echoing the illustrative example introduced earlier. Consistent with the literature on LLM-based research, we record the choice made by the LLM and treat it as a proxy for human choice. Departing from prior work, we also record the model's belief—its internal probability distribution over the choice set—when a choice is generated. We then test whether model belief outperforms model choice itself in estimating ground-truth model choice in sample and predicting model choice out of sample, which serves as an arguably high-power test of model belief's performance gain over model choice. We present the implementation details below.

*Study Design and Data Collection*

We ask the LLM to make a diaper brand choice. The diaper category is frequently examined in the marketing literature to study consumer preferences and choices (e.g., [Lin, Zhang, and Hauser](#)

2015; Tehrani and Ching 2024). It is also a common product category, one for which LLMs are likely to have rich contextual information. In our study, the LLM is asked to choose between two leading diaper brands, Pampers and Huggies, or an outside option (i.e., choosing neither brand). The outside option is included following established practice in the discrete choice modeling literature (Chib, Seetharaman, and Strijnev 2004).

To examine how the LLM's response changes with price, we vary the price of one brand while holding the price of the other brand constant. We fix the unit price of Huggies at 30 cents and systematically vary the unit price of Pampers from 25 cents to 40 cents in 1-cent increments, creating 16 choice scenarios, each corresponding to a Pampers price point. These price levels are chosen based on responses from the LLM used in this study regarding the typical prices of Huggies and Pampers diapers. This approach helps ensure that the choice task appears "natural" relative to the LLM's contextual knowledge. The chosen price levels are also consistent with industry reports.[3]

Although not central to our study, it is worth noting that each run is independent and the LLM does not "know" its outputs will be aggregated or its responses will be contrasted across price points. This feature mitigates concerns that the number of attribute levels might alter the measured importance of that attribute (Wittink, Krishnamurthi, and Reibstein 1990) or that respondents might feel compelled to vary their answers in within-subject designs (Charness, Gneezy, and Kuhn 2012).[4]

Except for the Pampers price, the rest of the input prompt remains constant across all choice scenarios (see the Appendix for the prompt). To elicit relevant model responses, the prompt informs the LLM that "You are visiting a store to buy baby diapers." The prompt also includes descriptions of Pampers and Huggies diapers to contextualize the task. Without such context, the LLM may default to a simple numerical comparison of diaper prices, a tendency we confirm in a side test.

For a given input prompt, we query the LLM for its response, including its model belief and model choice. The LLM used for this study is OpenAI GPT-4o, which represents one of the most ad-vanced, publicly available models at the time of this research. In the Appendix, we provide a sample code snippet demonstrating how to query the OpenAI Chat Completions Application Programming Interface (API). In particular, we choose `temperature=1.0` to implement unbiased sampling. We extract model choice from the model's textual output. Importantly, we set `logprobs=True` to instruct the API to return the token-level log-probabilities and use the `logprobs.content` attribute of the model's response to extract the log-probability data, based on which we compute model belief over the choice alternatives. These simple code modifications allow researchers to collect model belief data at scale.

---

[3] See, for example, https://www.crossrivertherapy.com/research/diaper-facts-statistics, which reports that the "average cost of a disposable diaper in the U.S. is $.29."

[4] Blinded designs may introduce prompt ambiguity (Gui and Toubia 2025). We mitigate this concern by providing the competitor's price and detailed context. We focus on comparing model belief and model choice under a common prompt, leaving prompt optimization to future research.

For each unique input prompt, which corresponds to a unique choice scenario defined by the Pampers price, we run the API call 1,000 times. These repeated runs aim to approximate the asymptotic choice probabilities, which serve as a proxy for the "ground truth" market share within this simulated environment. Taken together, this study design involves a total of 16 choice scenarios × 1,000 runs = 16,000 runs.

Two comments on the study design are in order. First, we aim to demonstrate the performance difference between model belief and model choice in the simplest possible setting. Therefore, we deliberately present a stylized choice task in response to price variations alone, holding other product attributes constant. We also focus on a representative consumer in the baby diaper market, as emulated by the LLM, while abstracting away from consumer heterogeneity. Nevertheless, we expect the insight to generalize to settings where the LLM responds to different price-attribute combinations, assuming the role of various consumer personas. The key insight is that model belief captures not only how various decision factors influence the eventual choice, but also how they shape the likelihood of making a choice at all.

Second, we treat LLM-generated choices as the ground truth to evaluate the performance of model belief versus model choice. While choices in actual markets may or may not differ from those generated by the LLM, the primary goal of this empirical study is to show that model belief contains richer information and provides a more accurate measure of demand than model choice, all within this self-contained LLM-based system. As explained, we test this claim by examining whether model belief outperforms model choice in explaining and predicting the model's own choices. The alignment between LLM-simulated behavior and real-world consumer behavior, as well as prompt design strategies to achieve such alignment, has been extensively studied in the literature and is not the primary target of this research.

*Estimation Method*

We use the multinomial logit (MNL) model to estimate the demand curve for Pampers, the focal brand in this study. The MNL is one of the most frequently adopted discrete choice models (Guadagni and Little 1983; Train 2009). Despite its simplicity, it effectively captures and predicts consumer choice among multiple alternatives, the choice context addressed in this study.

Specifically, let $j \in \mathcal{J} \equiv \{Pampers, Huggies, neither\}$ index the available alternatives and $s$ index the $S = 16$ choice scenarios corresponding to the 16 Pampers price points. Let $p_{js}$ denote the price of alternative $j$ in choice scenario $s$, which varies across choice scenarios for Pampers, is fixed at 30 cents for Huggies, and is normalized to 0 for the "neither" option. The utility that a consumer derives from choosing alternative $j$ in choice scenario $s$ is given by:

$$u_{js} = \alpha_j + \beta \times p_{js} + \epsilon_{js}$$

where $\alpha_j$ is a brand-specific intercept for alternative $j$ and $\beta$ denotes price sensitivity. The term $\epsilon_{js}$ is the idiosyncratic error, assumed to be independently and identically distributed according to a Type-I extreme value distribution. Because only utility differences matter in MNL, the utility of the outside option "neither" is normalized to 0. Thus, the utility $u_{js}$ for either brand is the utility relative to the outside option.

The probability that the consumer chooses alternative $j$ in choice scenario $s$, denoted by $P_{js}$, follows the standard logit formula:

$$P_{js} = \frac{\exp u_{js}}{\sum_{j' \in \mathcal{J}} \exp u_{j's}}.$$

The vector of parameters to estimate is $\theta = (\alpha_{\text{Pampers}}, \alpha_{\text{Huggies}}, \beta)$. Maximum likelihood estimation (MLE) yields consistent estimates of $\theta$, where the log likelihood function to maximize is:

$$\mathcal{L}(\theta) = \sum_{j \in \mathcal{J}} \sum_{s=1}^{S} d_{js} \log P_{js}(\theta).$$

In standard MNL, $d_{js}$ denotes model choice, which equals 1 if alternative $j$ is chosen in choice scenario $s$ and 0 otherwise. The resulting MLE parameter estimates of $\theta$ are a smooth function of the choice data. Therefore, based on Proposition 3, maximizing the same log likelihood function but replacing $d_{js}$ with model belief also yields consistent estimates of $\theta$. As discussed, this is one of the useful features of the model belief measure. The estimation procedure for model belief and model choice thus only differs in which model response measure is used for $d_{js}$.

We randomly split the full sample into a "training set" for parameter estimation and a holdout "test set" for evaluating out-of-sample predictions. To prevent data leakage (Ludwig, Mullainathan, and Rambachan 2025), we split the full sample by Pampers price, so that prices in the test set are never present in the training set. Specifically, the test set consists of three choice scenarios that correspond to three randomly chosen Pampers prices (28, 31, and 37) while the training set contains the remaining 13 choice scenarios. This results in an approximately 80-20 split between estimation and test samples.

*Results*

**LLM Responses**

We begin by describing the LLM's raw response data. Table 1 presents the top 10 model response patterns by frequency across the 16 choice scenarios and 1,000 runs per choice scenario. Each response pattern contains a unique sequence of tokens up to the pivot token, which unambiguously indicates the model's choice in that run. (A single response pattern may correspond to multiple responses that differ in the sequence of tokens following the pivot position.) These top 10 response patterns account for around 95% of all responses.

**Table 1:** Top 10 LLM Response Patterns in This Study.

| LLM Response Pattern (Terminated at Pivot Token) | Count | Frequency |
|---|---|---|
| I would choose ** P | 5,934 | 37.09% |
| I would choose ** H | 4,244 | 26.52% |
| I would choose to buy ** P | 3,664 | 22.90% |
| I would choose to buy ** H | 475 | 2.97% |
| Based on the information provided , I would choose ** P | 269 | 1.68% |
| I would buy ** P | 151 | .94% |
| I would choose Pamp | 151 | .94% |
| Based on the information provided , I would choose to buy ** P | 121 | .76% |
| I would buy ** H | 96 | .60% |
| Based on the information provided , I would choose ** H | 86 | .54% |

*Notes:* The full sample of LLM responses in this study contains 16,000 responses, corresponding to the 16 choice scenarios (as determined by Pampers price) and 1,000 runs per choice scenario. The response patterns are recorded verbatim, with spacing around punctuation marks deliberately preserved and the ** symbol denoting the Markdown syntax used by the OpenAI API for bold text.

Once the pivot token is detected, the model choice data take the familiar form of discrete choices among Pampers, Huggies, or neither. The model belief data are computed from the LLM's logit distribution at the pivot token. The position of the pivot token varies across prompts and runs, with a mean of 6.29 and a standard deviation of 4.81 across all responses. As discussed earlier in the paper, variation in LLM response sequences and pivot positions makes the asymptotic equivalence between model belief and mean model choice a nontrivial property.

**Ground-Truth Demand Curve Based on LLM Responses**

We construct the ground-truth demand curve based on the full set (including both estimation and test sets) of actual model choice data at each Pampers price point. Figure 1 presents the results. The left panel reports the market share of Pampers at each price, averaged across all 1,000 runs for that price. The gray solid line is the ground-truth demand curve based on actual model choice data. The black dashed line is the demand curve based on actual model belief data, where model belief substitutes for model choice in determining the market share of each alternative. As discussed, market shares are smooth functions of choice distributions, hence the market shares based on model choice and model belief are asymptotically equivalent according to Proposition 3. Indeed, the two demand curves are almost indistinguishable.

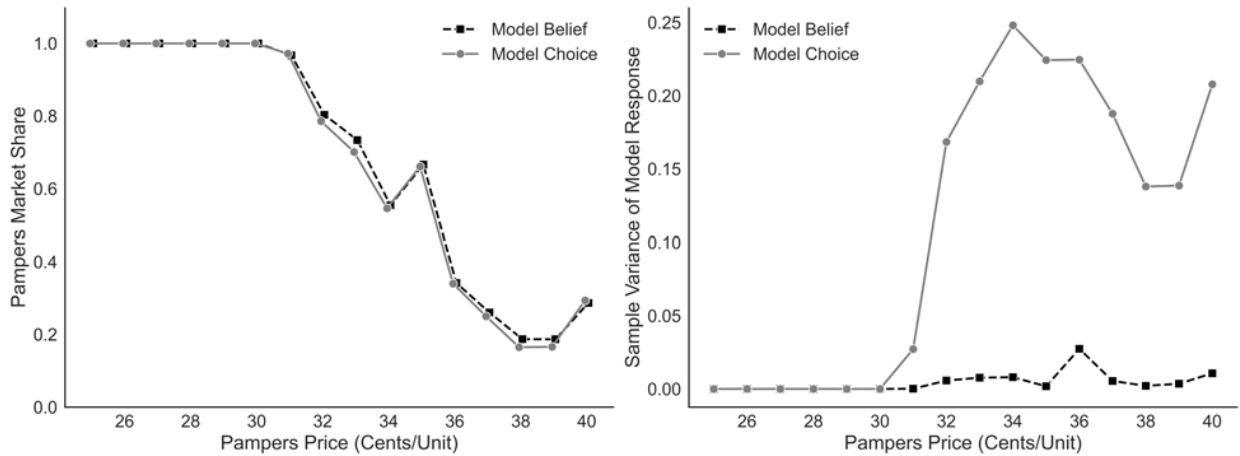

**Figure 1:** Ground-Truth Demand Curve Based on LLM Responses.

*Notes:* The left panel shows the average market share of Pampers across all 1,000 runs at each Pampers price based on model belief and model choice, respectively. The two curves are nearly identical but slightly jittered horizontally to ensure visual distinction in the figure. The right panel shows the sample variance of model belief and model choice, respectively, across the 1,000 runs for each Pampers price.

The demand curve itself behaves plausibly: overall, market share tends to decline as price rises, consistent with economic intuition. This result reinforces the view that LLMs have the potential to serve as proxies for human respondents in market research (e.g., Arora, Chakraborty, and Nishimura 2025; Brand, Israeli, and Ngwe 2024). Below 30 cents, the competing brand Huggies' price, Pampers is almost always chosen. Beyond this price point, Pampers demand overall declines with price and remains positive across the range of prices examined in this study. These results are consistent with Pampers being a premium brand in the diaper category, with both a higher average price and a higher market share than Huggies (Tehrani and Ching 2024). Moreover, the LLM in this study is prompted to be visiting a store to buy diapers, which may explain the high demand for diapers overall. There are upticks in the demand curve at prices 35 and 40. We leave it to future research to ascertain the reason, but one possible explanation is that rounded numbers appear more frequently in written languages and may be processed differently by humans (e.g., Wadhwa and Zhang 2015) and potentially by LLMs.

The right panel of Figure 1 further examines the empirical variation in model responses underlying the demand curve. For each Pampers price, the figure plots the sample variance of model belief versus model choice for Pampers across the 1,000 runs at that price. As predicted by Proposition 2, model choice shows substantially higher variance than model belief. For a rough calculation, we average the sample variance at each Pampers price (an unbiased estimator of the true variance at that price) across all prices, which yields .0046 for model belief and .1109 for model choice. Based on these numbers and Proposition 2, achieving the same level of accuracy with the model choice estimator would require approximately 24 times as many runs as with model belief. (The ratio for downstream quantities depends on how these quantities are constructed, according to Proposition 4. We examine this ratio empirically later.) At a 95% confidence level ($\delta = .05$) and with 1,000 runs, these sample variance values correspond to an accuracy tolerance of $\epsilon = .0096$ for model belief and $\epsilon = .0471$ for model choice. In terms of effect size, these accuracy tolerance levels in turn correspond to 1.41% deviation for model belief from the average Pampers market share in the full sample (68.01%) and 6.93% deviation for model choice, suggesting that 1,000 runs approximate asymptotic model responses with reasonable accuracy.

Substantively, both model belief and model choice show little variation below the competitor's price of 30 cents. This result echoes the left panel of Figure 1 and suggests that, in the data, Pampers is a "no-brainer" choice at these low prices. Beyond 30 cents, both measures exhibit higher sample variance. At 34 cents, model choice reaches its highest variance while model belief maintains a much lower variance. This is the price point corresponding to about 50% Pampers market share. Intuitively, this may indicate the point around which the most brand switching occurs and the most consumers are near indifference in their choice. While model choice depicts switching around this price as an abrupt shift in decisions, model belief captures the more subtle, gradual change in underlying preferences.

**Effect of Temperature on LLM Responses**

We have set an LLM temperature of 1 to implement unbiased sampling. For completeness, we present how different temperature values affect model responses. Figure 4 in the Appendix shows the results across the 1,000 runs when Pampers price is 31. This price is chosen for illustration because the corresponding market shares are neither so extreme as to conceal differences between model belief and model choice, nor so balanced as to obfuscate the LLM's tendency toward giving even answers under high temperatures.

At a temperature of 1, the Pampers market share exhibits nearly identical mean values across the 1,000 runs for model belief and model choice, but a much larger standard deviation for model choice, consistent with Propositions 1 and 2. For temperature values below 1, model choice tends to overrepresent the dominant option (Pampers at 31 cents), as expected, whereas model belief tends to reflect the ground-truth Pampers market share at that price. As the temperature increases past 1, market shares become more balanced under both measures, although the trend is more pronounced under model choice. A useful implication for LLM-based research, as a side note, is that while temperature settings can be used to modulate LLM response randomness, they may introduce bias into the average response.

Importantly, when the temperature value differs from 1, there is no guarantee that model belief will converge to its ground-truth level. This may appear counterintuitive at first glance because temperature only affects the sampling of tokens without altering their underlying probability distribution. However, the temperature value influences token sampling from the very first position in the output sequence, which in turn shapes the distribution of subsequent tokens and ultimately determines the pivot token and its associated model belief. As discussed, this sequential sampling process, together with the dependence of model belief on the pivot token, makes the asymptotic equivalence between model belief and mean model choice a nontrivial result.[5]

**Estimation Results and Predictive Performance**

We now turn to the comparison between model belief and model choice in their demand estimation performance. In particular, we examine the claim, presented with the opening example, that model belief provides more accurate estimates of price sensitivity and predicts out-of-sample choices better than model choice itself. To do so, we estimate the choice model specified earlier in the paper on the training set (of 13 choice scenarios) and examine its predictive performance on the test set (of the remaining 3 choice scenarios).

---

[5]Even at a temperature of 0, model belief varies across runs for the same input prompt. This may be related to the internal workings of the OpenAI API, which are beyond the scope of this paper.

We first estimate the choice model using the ground-truth model choice data in the full training set of $13 \times 1,000$ runs. The parameter estimates serve as the ground-truth benchmark for subsequent comparisons. We similarly estimate the choice model using the model belief data in this full training set. According to Proposition 3, the two sets of estimates should be similar, as 1,000 runs per choice scenario should reasonably approximate the asymptotic behavior of either model response.

We re-estimate the choice model using either measure of model response, varying the number of runs per choice scenario from 1 to 1,000. For example, to generate estimation results in the 1-run case, we randomly draw 1 run of model response data for each of the 13 choice scenarios from the full training set and use the drawn data for estimation. According to Proposition 4, model belief should yield more precise estimates than model choice in finite numbers of runs.

For a robust performance assessment, we repeat the aforementioned random-draw process with replacement 1,000 times (Efron 1992). For example, in the case of 1 run per choice scenario, we randomly draw 1 run from that choice scenario's 1,000 runs in the full training set, repeating this procedure 1,000 times with replacement. This approach allows us to compare the average performance of model belief and model choice across a large number of random draws, so that the conclusion is not driven by the realization of a particular draw.

The upper panel of Table 2 presents the estimation results on the training set for 1,000, 1, 2, 3, and 100 runs per choice scenario, respectively. (We feature these run counts for their informational value. Overall, parameter estimates change at a diminishing rate as the number of runs increases; the changes are most pronounced over the first few runs and almost invisible past 100 runs.) For each parameter in $\theta$, we report its point estimate and standard error (SE), both averaged across the 1,000 random draws, as well as the associated $p$-value. The point estimate, when compared against its ground-truth value, reflects the accuracy of the parameter estimate, whereas the standard error captures its precision. We also report the standard deviation (SD) of the point estimate across the 1,000 random draws, which indicates the robustness of the estimate to sampling variability.

**Table 2:** Estimation Results and Predictive Performance

| Number of Runs per Scenario | 1,000 Runs | | 1 Run | | 2 Runs | | 3 Runs | | 100 Runs | |
|---|---|---|---|---|---|---|---|---|---|---|
| Model Response Measure | Model Choice (Ground Truth) | Model Belief | Model Choice | Model Belief | Model Choice | Model Belief | Model Choice | Model Belief | Model Choice | Model Belief |
| **Estimation Results on the Training Set** | | | | | | | | | | |
| Price coefficient | -.463 | -.463 | -8.754 | -.468 | -.846 | -.466 | -.549 | -.465 | -.464 | -.463 |
| Price coefficient SE | (.008) | (.008) | (138.090) | (.264) | (6.193) | (.186) | (1.421) | (.151) | (.026) | (.026) |
| Price coefficient $p$-value | .000 | .000 | .949 | .077 | .891 | .012 | .699 | .002 | .000 | .000 |
| Price coefficient sample SD | [.000] | [.000] | [19.488] | [.046] | [2.522] | [.035] | [.533] | [.027] | [.022] | [.005] |
| | | | | | | | | | | |
| Pampers intercept | 32.016 | 31.989 | 348.830 | 32.250 | 46.531 | 32.183 | 35.184 | 32.250 | 31.982 | 32.159 |
| Pampers intercept SE | (47.443) | (46.188) | (62,749.002) | (1,673.434) | (4,222.442) | (1,080.319) | (2,173.329) | (995.442) | (146.164) | (182.623) |
| Pampers intercept $p$-value | .500 | .489 | .996 | .985 | .991 | .976 | .987 | .974 | .827 | .860 |
| Pampers intercept sample SD | [1.497] | [1.500] | [735.594] | [2.267] | [96.704] | [1.918] | [19.427] | [1.848] | [1.693] | [1.600] |
| | | | | | | | | | | |
| Huggies intercept | 29.444 | 29.346 | 305.534 | 29.584 | 42.246 | 29.527 | 32.190 | 29.596 | 29.403 | 29.516 |
| Huggies intercept SE | (47.443) | (46.187) | (62,688.188) | (1,673.419) | (4,244.228) | (1,080.309) | (2,177.166) | (995.434) | (146.163) | (182.622) |
| Huggies intercept $p$-value | .535 | .525 | .996 | .986 | .992 | .978 | .988 | .976 | .841 | .872 |
| Huggies intercept sample SD | [1.497] | [1.500] | [641.020] | [2.155] | [86.052] | [1.842] | [17.275] | [1.804] | [1.651] | [1.599] |
| | | | | | | | | | | |
| Log likelihood | -4,998.315 | -4,965.936 | -3.744 | -4.938 | -9.026 | -9.900 | -13.996 | -14.862 | -499.298 | -496.466 |
| Number of observations | 39,000 | 39,000 | 39 | 39 | 78 | 78 | 117 | 117 | 3,900 | 3,900 |
| **Predictive Performance on the Test Set** | | | | | | | | | | |
| RMSE | .003 | .004 | .016 | .004 | .010 | .004 | .007 | .004 | .004 | .004 |
| RMSE_Diff | - | .000 | .013 | .001 | .007 | .001 | .004 | .001 | .000 | .000 |
| RMSE_Diff equality test $t$-stat. | - | | 22.264 | | 18.866 | | 13.364 | | -8.646 | |
| RMSE_Diff equality test $p$-value | - | | .000 | | .000 | | .000 | | 1.000 | |
| | | | | | | | | | | |
| MAE | .044 | .046 | .070 | .046 | .057 | .046 | .050 | .046 | .044 | .046 |
| MAE_Diff | - | .002 | .026 | .002 | .013 | .001 | .006 | .002 | -.000 | .002 |
| MAE_Diff equality test $t$-stat. | - | | 23.223 | | 12.317 | | 5.190 | | -7.588 | |
| MAE_Diff equality test $p$-value | - | | .000 | | .000 | | .000 | | 1.000 | |

*Notes:* The training set comprises 13 choice scenarios, and the test set comprises another 3. Predictive performances for all choice models are evaluated on the same ground-truth test set. SE is the standard error of the parameter estimate. SD is the standard deviation of the point estimate across the 1,000 random draws. RMSE_Diff and MAE_Diff are the difference in RMSE and MAE from their respective values in the ground-truth case.

In the ground-truth case, price sensitivity is precisely estimated to be $-.463$ (SE $=.008$, $p=.000$). Pampers exhibits a higher brand intercept than Huggies, but neither parameter is precisely estimated. One possible explanation, which future research may examine, is that the prompt used for this study does not elicit clear enough brand preferences from the LLM. The parameter estimates exhibit indiscernible or small standard deviations across the random draws. This is expected because the full training set is sampled in the ground-truth case. The small deviations arise because the initial values of the parameters during estimation are randomly set in each draw, a practice intended to yield more robust estimates.

Estimation results based on model belief over 1,000 runs per choice scenario are statistically indistinguishable from their ground-truth counterpart. The price sensitivity estimates are nearly identical between model belief and model choice ($p=.988$ for a two-tailed test of equality). This finding is consistent with the asymptotic equivalence between downstream smooth functions of model belief and model choice, as stated in Proposition 3.

With only 1 run of model response data per choice scenario, estimation results become less accurate, precise, and robust, as expected. However, model belief performs substantially better than model choice on all three criteria. In particular, the price sensitivity estimate of $-.468$ based on model belief is not far from the ground-truth value of $-.463$, is estimated with marginal precision (SE $=.264$, $p=.077$), and is relatively robust with a standard deviation (SD $=.046$) about 1/10th of the point estimate. Model choice yields a price sensitivity of $-8.754$ that is imprecisely estimated (SE $=138.090$, $p=.949$) and less robust to sampling variability (SD $=19.488$). We do not compare model belief and model choice based on log likelihood, which is not directly comparable between the two model response measures in finite samples.[6]

As the run count increases to 2, 3, and 100 per choice scenario, both model belief and model choice improve in the accuracy, precision, and robustness of their parameter estimates. Model belief continues to outperform model choice on all three criteria, although the performance gap narrows as the number of runs grows. In the 100-run case, price sensitivity is precisely estimated to be around its ground-truth value for both model belief and model choice (SE = .026, $p$ = .000), and its difference between these two model response measures is almost 0 ($p$ = .978 for a two-tailed test of equality).

---

[6]Rarely chosen options produce extremely low log-likelihood values. While a model choice of 0 removes these values from the total log-likelihood calculation, model beliefs close to 0, although potentially more informative, permit such values to diminish the total log likelihood.

Next, we evaluate the predictive performance of model belief relative to model choice. For a fair comparison, we evaluate all choice models, regardless of response measure or number of runs, on the same ground-truth test set of model choice data in the 3 holdout choice scenarios across 1,000 runs per scenario. We examine two standard predictive-accuracy metrics: root mean squared error (RMSE) and mean absolute error (MAE), noting that MAE is less sensitive to outliers.

The lower panel of Table 2 shows the predictive performance results across the 1,000 random draws. As a benchmark, model choice over 1,000 runs per choice scenario (the ground-truth case) yields an RMSE of .003 and an MAE of .044. In comparison, model belief attains an RMSE of .004 and an MAE of .046 even with a single run, and both metrics remain essentially unchanged as the run count increases. Model choice, on the other hand, produces an RMSE of .016 and an MAE of .070 with 1 run, although both metrics decline with more runs.

For a more formal test, we compute RMSE_Diff, the difference in RMSE between each choice model and the ground-truth benchmark, where a larger value means relatively worse predictive accuracy. We compute MAE_Diff similarly. For a given run count, the null hypothesis is that model choice has the same RMSE_Diff and MAE_Diff as model belief across the 1,000 random draws, whereas the alternative hypothesis is for model choice to have a larger RMSE_Diff and a larger MAE_Diff. As the lower panel of Table 2 shows, the null hypothesis is strongly rejected ($p$ = .000 for a one-tailed test of equality) in the cases of 1, 2, and 3 runs, and is not rejected at 100 runs ($p$ = 1.000 for a one-tailed test of equality). That is, model belief predicts out-of-sample model choice significantly better than model choice itself for limited numbers of runs, although this predictive advantage diminishes as the run count increases.

*Discussion*

We have shown that model belief yields more accurate, precise, and robust price sensitivity estimates and predicts choices better than model choice itself, especially in small numbers of runs. We now revisit the intuition behind the better performance of model belief. We visualize the intuition with Figure 2, where we plot the estimated demand curves based on parameter estimates from model belief versus model choice. In all four panels corresponding to 1, 2, 3, and 100 runs per choice scenario, respectively, we also plot the estimated demand curves based on ground-truth parameter estimates. All curves are averaged across the 1,000 random draws for a robust assessment.

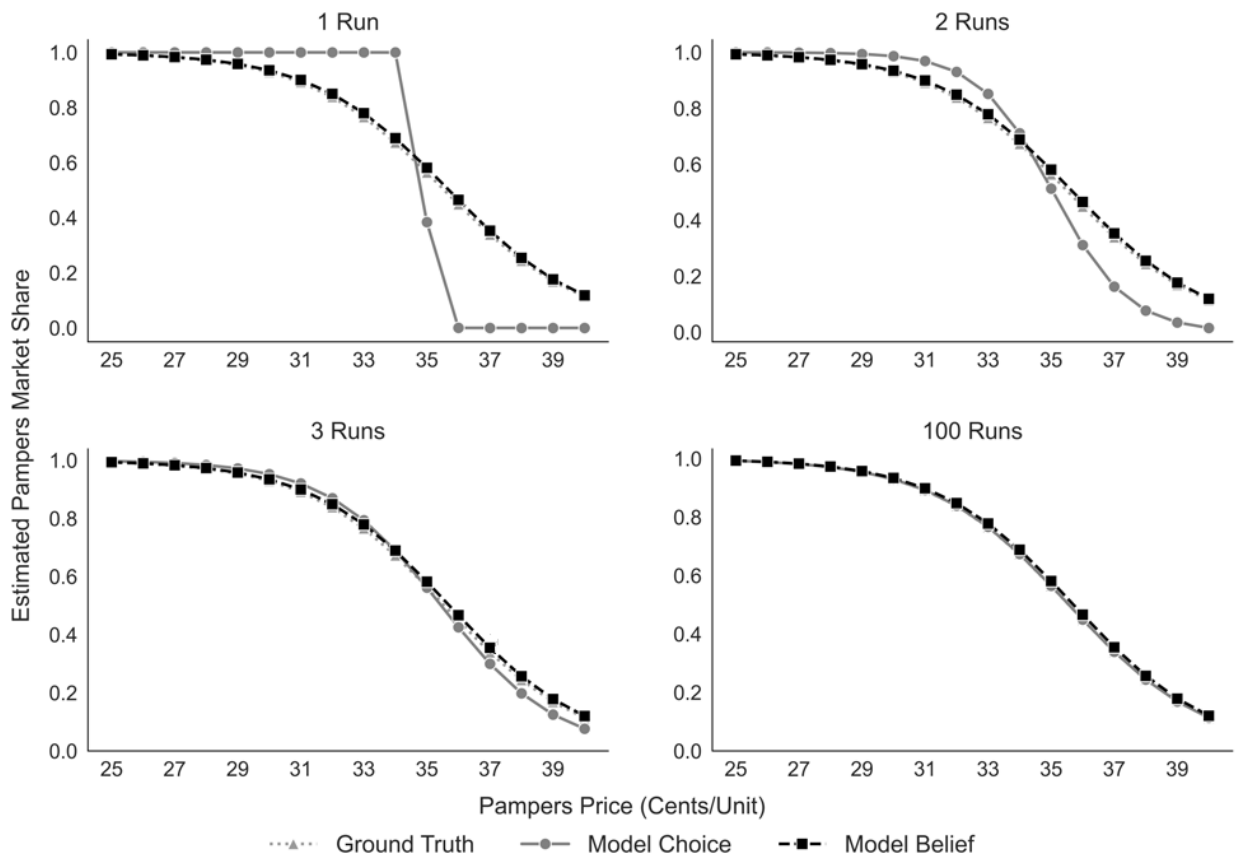

**Figure 2:** Estimated Demand Curves.

*Notes:* This figure presents the demand curves estimated from model belief versus model choice for 1, 2, 3, and 100 runs per choice scenario, respectively. Each curve is based on the average across the 1,000 random draws.

In all four panels, the demand curve estimated from model belief is nearly indistinguishable from the curve estimated from the ground-truth data. The demand curve estimated from model choice visibly deviates from both, although the gap narrows as the run count increases. Importantly, for small run counts, model choice overstates price sensitivity in the middle region of prices and understates it in the tail regions. This result is consistent with the intuition described in the opening example. In the mid-price range where price variations are more likely to trigger switching, model choice captures switching as discrete shifts in demand, whereas model belief reflects a more gradual change in the underlying preference. In the price tails, very high or very low, equally sized price changes may fail to flip brand choice, even if they meaningfully shift the underlying preference, an effect that model belief captures but model choice can miss.

A natural question that remains is how many runs are "good enough" for model belief versus model choice. Propositions 2 and 4 theoretically prove that, for any target accuracy and confidence levels, the number is weakly smaller for model belief, and strictly smaller unless choices are deterministic. However, the exact run count that guarantees acceptable accuracy for downstream quantities, such as price sensitivity, may not have simple analytical expressions. Therefore, we approach this question empirically with bootstrapping.

We focus this analysis on price sensitivity, the parameter of interest. We set the target accuracy tolerance $\epsilon$ to be a percentage of the ground-truth price sensitivity in absolute value. Each choice model is estimated 1,000 times on randomly drawn model response data in the training set. This procedure approximates, for each choice model, the probability of its price sensitivity estimate falling within the target percentage around its ground-truth value.

Figure 3 presents the results for accuracy tolerances that are 10% and 5%, respectively, of the ground-truth price sensitivity. For both accuracy criteria, the probability of obtaining accurate price sensitivity estimates increases with run count for both model belief and model choice, as expected. However, model belief attains the same level of accuracy with substantially fewer runs. At the standard 95% confidence level ($\delta = .05$), model belief requires 5 runs to reach 10% accuracy tolerance compared with 93 for model choice, and 16 runs to reach 5% tolerance compared to 383. These numbers translate to a reduction of run count by a factor of 19 and 24, respectively, when model belief is used instead of model choice.

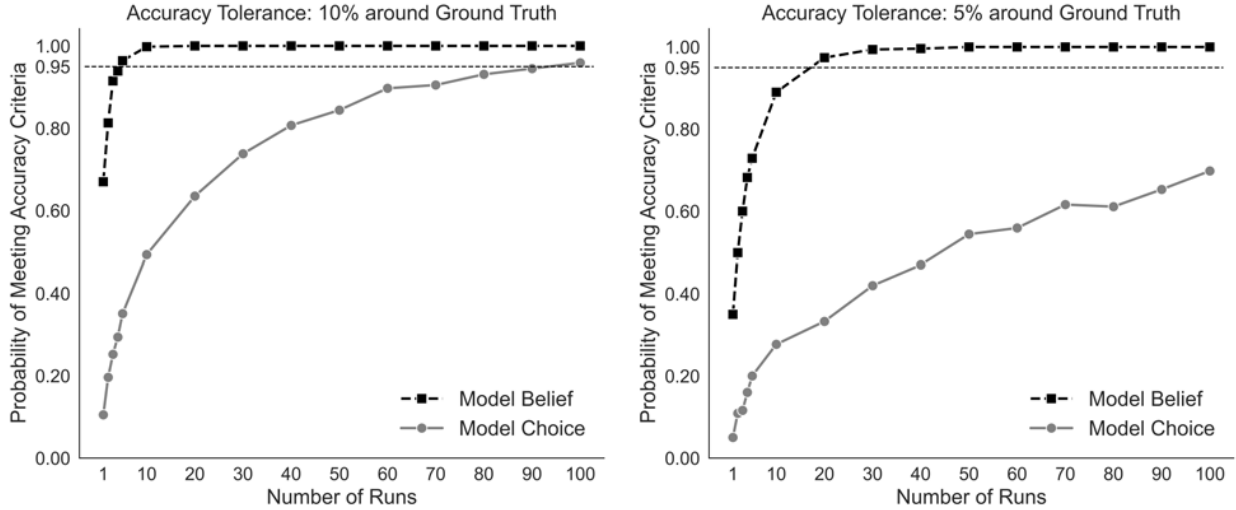

**Figure 3:** Price Coefficient Accuracy by Number of Runs.

*Notes:* This figure presents the probability of the estimated price coefficient falling within 10% (left) or 5% (right) around the ground-truth value as the number of runs increases. At the 95% confidence level, model belief requires 5 runs per choice scenario to reach 10% accuracy tolerance compared with 93 for model choice (left), and 16 runs to reach 5% tolerance compared to 383 (right).

Model belief's much lower run-count requirement can greatly lower the cost of LLM-based research. The proportional reduction in run count can mean substantial savings in computational expenses in more complex applications. Moreover, the sharply reduced computational time required by model belief not only accelerates discovery but can also be essential in latency-critical settings such as real-time price promotions. Last but not least, the heavy computational demands of LLMs have raised environmental, social, and economic concerns about their sustainability. Model belief, as a higher insight-per-compute alternative to model choice, suggests a meaningful step toward sustainable computing.

## TOWARD A NEW MEASURE FOR LLM-BASED RESEARCH

This paper introduces and validates model belief as a theoretically principled and empirically superior measure for extracting the latent knowledge of LLMs. Our analysis begins with the premise that when obtaining responses, whether from human or silicon subjects, it is often more valuable to understand the preferences among possible options, not merely to note a single observed choice. In human-based studies, this task can be constrained by the difficulty of eliciting precise quantification of latent preferences. LLMs offer a powerful solution by enabling direct access to the internal preference distribution underlying any manifested choice—information that our proposed model belief measure succinctly captures.

We formally define model belief as an LLM's softmax probability distribution over choice alternatives at the pivot token, where the model choice is ultimately made, in each generation run. We prove that model belief is asymptotically equivalent to the mean of model choices, while providing a more efficient estimator with lower variance and faster convergence. (Here, asymptotic equivalence is nontrivial because model belief depends on the pivot token, which may vary across runs.) We also prove that analogous properties hold for downstream quantities that are smooth functions of model belief and model choice. These theoretical properties matter because brute-force resampling—the classical way to recover an LLM's choice distribution by querying it for model choices repeatedly—can be computationally expensive, if not prohibitive.

We examine the empirical performance of model belief in an LLM-based demand estimation study. In finite samples typical of practical applications, model belief produces more accurate, precise, and robust estimates of price sensitivity than model choice, and predicts model choices out of sample significantly better than model choice itself. Achieving the same accuracy using model choice would require roughly 20 times as many runs as with model belief in this setting. While the performance gap between the two measures narrows as run count increases, this convergence is slow and can be impractical to attain. These empirical findings, together with the theoretical properties, establish the advantage of model belief as a new standard measure: in LLM-based research, model belief should serve as the primary metric of LLM response.

While this paper focuses on the use case of demand estimation, the benefits of model belief extend to many other domains. Any research domain seeking to elicit preferences, distributional knowledge, or probabilistic forecasts from an LLM stands to benefit from the richer information contained in model belief. The scope of measurement tasks enhanced by model belief also extends beyond parameter estimation. In what follows, we highlight several other applications.

First, model belief directly unveils the choice alternatives an LLM may consider. This informa-tion can help researchers uncover the consideration set of the LLM, or of the human decision-makers the LLM is trained to represent. In the marketing literature, it is well-established that the decision to consider is distinctively informative (Hauser and Wernerfelt 1990) and consideration information is critical for understanding consumer preferences (Bronnenberg and Vanhonacker 1996). Yet the consideration set is often unobserved and must be elicited from consumers or inferred from choice data probabilistically or as functions of consumer and firm actions (e.g., Ben-Akiva and Boccara 1995; Mehta, Rajiv, and Srinivasan 2003; Roberts and Lattin 1991). Model belief provides a new source of information for identifying consideration.

Second, in studies concerned with rare events, model belief offers a unique advantage. By providing the full probability distribution over potential choices, it allows researchers to efficiently identify low-probability scenarios that would be especially difficult to detect through repeated sampling of model choice. This capability of model belief can substantially enhance rarity-based applications—such as recognizing emerging phenomena or ideas, preserving minority views, detecting early warnings, assessing tail risks, or completing scenario plans—applications that are otherwise practically challenged with conventional resampling methods. Relatedly, model belief provides a more efficient metric than model choice for uncertainty-sensitive applications, such as financial planning under varying risk preferences and inventory planning with volatile demand.

Third, the finer granularity of information in model belief, relative to model choice, can be particularly valuable in sequential decision settings. Sequential learning from coarsened information, even if rational at the level of the individual learner, can give rise to irrational information cascades or herding behaviors at the aggregate level—a failure of the wisdom of the crowds well-established both theoretically (Banerjee 1992; Bikhchandani, Hirshleifer, and Welch 1992) and empirically (Zhang 2010). This concern arises when LLM-generated data propagate across social networks, when LLMs are trained on data generated by LLMs (Shumailov et al. 2024), or when LLMs communicate with each other to perform various tasks. Transmitting model belief helps preserve the integrity of LLM outputs, whereas transmitting model choice remains susceptible to aggregate informational failure.

There are, however, practical caveats. Implementing model belief may be less straightforward with models that engage in complex, multi-step reasoning (e.g., chain-of-thought), such as OpenAI's o3 family. In these cases, two challenges emerge. First, the internal process leading to the final answer may not be fully accessible, especially when using API endpoints that only return the final output. Second, identifying the precise pivot token can be complicated by a model's tendency to weigh options or output ambiguous reasoning before producing a definitive answer.[7] These caveats do not undermine the theoretical validity of model belief, but they do highlight the need for future research on how to best extract and interpret belief distribution from various model architectures. They also underscore the need to specify which model outputs should be disclosed, in a way that balances confidentiality with the quality of downstream LLM applications.

To summarize, this work proposes a general principle for LLM-based research: whenever it adds value to recover an underlying distribution of options beyond the final choice, model belief should be the standard measure. Doing so not only enhances estimation accuracy and statistical efficiency but also ensures that research using LLMs fully exploits their distinctive affordances compared with human respondents. As LLM applications continue to evolve and proliferate across fields, model belief is poised to become a foundational tool for extracting richer information from LLMs at lower cost.

---

[7]Decision-first prompting can mitigate the latter challenge, but may overlook critical branching points in the reasoning process.

# REFERENCES

Aher, Gati V, Rosa I. Arriaga, and Adam Tauman Kalai (2023), "Using Large Language Models to Simulate Multiple Humans and Replicate Human Subject Studies," *Proceedings of the 40th International Conference on Machine Learning*, 202, 337–371.

Argyle, Lisa P, Ethan C Busby, Nancy Fulda, Joshua R Gubler, Christopher Rytting, and David Wingate (2023), "Out of One, Many: Using Language Models to Simulate Human Samples," *Political Analysis*, 31 (3), 337–351.

Arora, Neeraj, Ishita Chakraborty, and Yohei Nishimura (2025), "AI–Human Hybrids for Marketing Research: Leveraging Large Language Models (LLMs) as Collaborators," *Journal of Marketing*, 89 (2), 43–70.

Banerjee, Abhijit V (1992), "A Simple Model of Herd Behavior," *Quarterly Journal of Economics*, 107 (3), 797–817.

Ben-Akiva, Moshe and Bruno Boccara (1995), "Discrete Choice Models With Latent Choice Sets," *International Journal of Research in Marketing*, 12 (1), 9–24.

Bengio, Yoshua, Réjean Ducharme, Pascal Vincent, and Christian Jauvin (2003), "A Neural Probabilistic Language Model," *Journal of Machine Learning Research*, 3 (Feb), 1137–1155.

Bikhchandani, Sushil, David Hirshleifer, and Ivo Welch (1992), "A Theory of Fads, Fashion, Custom, and Cultural Change as Informational Cascades," *Journal of Political Economy*, 100 (5), 992–1026.

Blanchard, Simon J, Nofar Duani, Aaron M Garvey, Oded Netzer, and Travis Tae Oh (2025), "New Tools, New Rules: A Practical Guide to Effective and Responsible GenAI Use for Surveys and Experiments Research," *Journal of Marketing*, forthcoming.

Brand, James, Ayelet Israeli, and Donald Ngwe (2024), "Using LLMs for Market Research," *Harvard Business School Working Paper*.

Bronnenberg, Bart J and Wilfried R Vanhonacker (1996), "Limited Choice Sets, Local Price Response, and Implied Measures of Price Competition," *Journal of Marketing Research*, 33 (2), 163–173.

Charness, Gary, Uri Gneezy, and Michael A Kuhn (2012), "Experimental Methods: Between-Subject and Within-Subject Design," *Journal of Economic Behavior & Organization*, 81 (1), 1–8.

Chib, Siddhartha, PB Seetharaman, and Andrei Strijnev (2004), "Model of Brand Choice With a No-Purchase Option Calibrated to Scanner-Panel Data," *Journal of Marketing Research*, 41 (2), 184–196.

Chintagunta, Pradeep K and Harikesh S Nair (2011), "Discrete-Choice Models of Consumer Demand in Marketing," *Marketing Science*, 30 (6), 977–996.

De Freitas, Julian, Gideon Nave, and Stefano Puntoni (2025), "Ideation With Generative AI—in Consumer Research and Beyond," *Journal of Consumer Research*, 52 (1), 18–31.

Demszky, Dorottya, Diyi Yang, David S Yeager, Christopher J Bryan, Margarett Clapper, Susannah Chandhok, Johannes C Eichstaedt, Cameron Hecht, Jeremy Jamieson, Meghann Johnson et al. (2023), "Using Large Language Models in Psychology," *Nature Reviews Psychology*, 2 (11), 688–701.

Dillion, Danica, Niket Tandon, Yuling Gu, and Kurt Gray (2023), "Can AI Language Models Replace Human Participants?," *Trends in Cognitive Sciences*, 27 (7), 597–600.

Efron, Bradley (1992), "Bootstrap Methods: Another Look at the Jackknife," *Breakthroughs in Statistics: Methodology and Distribution*, Springer Series in Statistics, 569–593.

Goli, Ali and Amandeep Singh (2024), "Frontiers: Can Large Language Models Capture Human Preferences?," *Marketing Science*, 43 (4), 709–722.

Guadagni, Peter M and John DC Little (1983), "A Logit Model of Brand Choice Calibrated on Scanner Data," *Marketing Science*, 2 (3), 203–238.

Gui, George and Olivier Toubia (2025), "The Challenge of Using LLMs To Simulate Human Behavior: A Causal Inference Perspective," *arXiv preprint arXiv:2312.15524*.

Hauser, John R and Steven M Shugan (1980), "Intensity Measures of Consumer Preference," *Operations Research*, 28 (2), 278–320.

Hauser, John R and Birger Wernerfelt (1990), "An Evaluation Cost Model of Consideration Sets," *Journal of Consumer Research*, 16 (4), 393–408.

Horton, John J (2023), "Large Language Models as Simulated Economic Agents: What Can We Learn From Homo Silicus?," *National Bureau of Economic Research Working Paper 31122*.

Li, Peiyao, Noah Castelo, Zsolt Katona, and Miklos Sarvary (2024), "Frontiers: Determining the Validity of Large Language Models for Automated Perceptual Analysis," *Marketing Science*, 43 (2), 254–266.

Lin, Song, Juanjuan Zhang, and John R Hauser (2015), "Learning from Experience, Simply," *Marketing Science*, 34 (1), 1–19.

Ludwig, Jens, Sendhil Mullainathan, and Ashesh Rambachan (2025), "Large Language Models: An Applied Econometric Framework," *National Bureau of Economic Research Working Paper 33344*.

McFadden, Daniel (1986), "The Choice Theory Approach to Market Research," *Marketing Science*, 5 (4), 275–297.

Mehta, Nitin, Surendra Rajiv, and Kannan Srinivasan (2003), "Price Uncertainty and Consumer Search: A Structural Model of Consideration Set Formation," *Marketing Science*, 22 (1), 58–84.

Minaee, Shervin, Tomas Mikolov, Narjes Nikzad, Meysam Chenaghlu, Richard Socher, Xavier Amatriain, and Jianfeng Gao (2024), "Large Language Models: A Survey," *arXiv preprint arXiv:2402.06196*.

Roberts, John H and James M Lattin (1991), "Development and Testing of a Model of Consideration Set Composition," *Journal of Marketing Research*, 28 (4), 429–440.

Shumailov, Ilia, Zakhar Shumaylov, Yiren Zhao, Nicolas Papernot, Ross Anderson, and Yarin Gal (2024), "AI Models Collapse When Trained on Recursively Generated Data," *Nature*, 631 (8022), 755–759.

Tehrani, Shervin Shahrokhi and Andrew T Ching (2024), "A Heuristic Approach To Explore: The Value of Perfect Information," *Management Science*, 70 (5), 3200–3224.

Timoshenko, Artem, Chengfeng Mao, and John R Hauser (2025), "Can Large Language Models Extract Customer Needs as Well as Professional Analysts?," *arXiv preprint arXiv:2503.01870*.

Train, Kenneth E (2009), *Discrete Choice Methods with Simulation* Cambridge University Press.

Vaswani, Ashish, Noam Shazeer, Niki Parmar, Jakob Uszkoreit, Llion Jones, Aidan N Gomez, Łukasz Kaiser, and Illia Polosukhin (2017), "Attention Is All You Need," *Advances in Neural Information Processing Systems*, 30.

Wadhwa, Monica and Kuangjie Zhang (2015), "This Number Just Feels Right: The Impact of Roundedness of Price Numbers on Product Evaluations," *Journal of Consumer Research*, 41 (5), 1172–1185.

Wittink, Dick R, Lakshman Krishnamurthi, and David J Reibstein (1990), "The Effect of Differences in the Number of Attribute Levels on Conjoint Results," *Marketing Letters*, 1 (2), 113–123.

Ye, Zikun, Hema Yoganarasimhan, and Yufeng Zheng (2025), "LOLA: LLM-Assisted Online Learning Algorithm for Content Experiments," *Marketing Science*, forthcoming.

Zhang, Juanjuan (2010), "The Sound of Silence: Observational Learning in the U.S. Kidney Market," *Marketing Science*, 29 (2), 315–335.

## *APPENDIX*

### *Proof of Proposition 2*

Denote the variance of model belief and of model choice, respectively, as

$$V_b(j) := \text{Var}(b(j)) \quad \text{and} \quad V_c(j) := \text{Var}(\mathbb{1}\{c = j\}), \quad \forall j \in \mathcal{J}.$$

Let $Y_j = \mathbb{1}\{c = j\}$ be the random variable encoding whether model choice is $j$. By definition, $\text{Var}(Y_j) = V_c(j)$. Conditional on any given model belief $\boldsymbol{b}$, choice $c = j$ occurs with probability $b(j)$. Thus, the random variable $Y_j$ conditional on $\boldsymbol{b}$ follows a Bernoulli distribution: $Y_j \mid \boldsymbol{b} \sim \text{Bernoulli}(b(j))$. Hence,

$$\mathbb{E}[Y_j \mid \boldsymbol{b}] = b(j) \quad \text{and} \quad \text{Var}(Y_j \mid \boldsymbol{b}) = b(j)(1 - b(j)).$$

By the law of total variance:

$$V_c(j) = \text{Var}(Y_j) = \text{Var}(\mathbb{E}[Y_j \mid \boldsymbol{b}]) + \mathbb{E}[\text{Var}(Y_j \mid \boldsymbol{b})]$$
$$= \text{Var}(b(j)) + \mathbb{E}[b(j)(1 - b(j))]$$
$$= V_b(j) + \mathbb{E}[b(j)(1 - b(j))].$$

Since $b(j) \in [0, 1]$, the term $b(j)(1 - b(j)) \geq 0$, and therefore its expectation $\mathbb{E}[b(j)(1 - b(j))] \geq 0$. It follows that:

$$V_b(j) \leq V_c(j).$$

Here, equality holds if and only if $\mathbb{E}[b(j)(1 - b(j))] = 0$, which means $b(j)(1 - b(j)) = 0$ almost surely, i.e., $\Pr(b(j) \in \{0, 1\}) = 1$.

For $N$ independent runs,

$$\text{Var}\left(\hat{b}^{(N)}(j)\right) = \frac{V_b(j)}{N} \quad \text{and} \quad \text{Var}\left(\hat{c}^{(N)}(j)\right) = \frac{V_c(j)}{N}, \quad \forall j \in \mathcal{J},$$

which proves the first result in the proposition.

By Chebyshev's inequality,

$$N_b = \lceil \frac{V_b(j)}{\epsilon^2 \delta} \rceil \quad \text{and} \quad N_c = \lceil \frac{V_c(j)}{\epsilon^2 \delta} \rceil, \quad \forall j \in \mathcal{J},$$

which proves the second result in the proposition.

### *Proof of Proposition 4*

Denote the covariance of model belief and of model choice, respectively, as

$$\Sigma_b := \text{Var}(\boldsymbol{b}) \quad \text{and} \quad \Sigma_c := \text{Var}\big(\mathbb{1}\{c = j\}_{j=1}^{|\mathcal{J}|}\big).$$

By the central limit theorem, we have

$$\sqrt{N}\big(\hat{\boldsymbol{b}}^{(N)} - \bar{\boldsymbol{b}}\big) \xrightarrow{d} \mathcal{N}(0, \Sigma_b) \quad \text{and} \quad \sqrt{N}\big(\hat{\boldsymbol{c}}^{(N)} - \bar{\boldsymbol{c}}\big) \xrightarrow{d} \mathcal{N}(0, \Sigma_c).$$

Let $D_g$ be the $q \times |\mathcal{J}|$ Jacobian of $g$ evaluated at $\bar{\boldsymbol{c}}$. Then, by the delta method, the Gaussian limits for the plug-in estimators follow

$$\sqrt{N}\big(\hat{\boldsymbol{\theta}}_b^{(N)} - \bar{\boldsymbol{\theta}}\big) \xrightarrow{d} \mathcal{N}(0, D_g \Sigma_b D_g^\top) \quad \text{and} \quad \sqrt{N}\big(\hat{\boldsymbol{\theta}}_c^{(N)} - \bar{\boldsymbol{\theta}}\big) \xrightarrow{d} \mathcal{N}(0, D_g \Sigma_c D_g^\top).$$

We now prove the Loewner order of the model belief covariance and model choice covariance. Let $\boldsymbol{Y} = (\mathbb{1}\{c = 1\}, \cdots, \mathbb{1}\{c = |\mathcal{J}|\})$ be the one-hot encoded random vector for model choice. Conditional on any given model belief $\boldsymbol{b}$, the random vector $\boldsymbol{Y}$ follows a categorical distribution: $\boldsymbol{Y} \mid \boldsymbol{b} \sim \text{Categorical}(\boldsymbol{b})$. Hence,

$$\mathbb{E}[\boldsymbol{Y} \mid \boldsymbol{b}] = \boldsymbol{b} \quad \text{and} \quad \text{Var}(\boldsymbol{Y} \mid \boldsymbol{b}) = \text{Diag}(\boldsymbol{b}) - \boldsymbol{b}\boldsymbol{b}^\top.$$

By the law of total variance:

$$\Sigma_c = \text{Var}(Y) = \text{Var}(\mathbb{E}[Y \mid b]) + \mathbb{E}[\text{Var}(Y \mid b)]$$

$$= \text{Var}(b) + \mathbb{E}[\text{Diag}(b) - bb^\top]$$

$$= \Sigma_b + \mathbb{E}[\text{Diag}(b) - bb^\top].$$

For any given model belief vector $b$, the matrix $\text{Diag}(b) - bb^\top$ is the covariance matrix of a Categorical($b$) distribution, which is positive semidefinite. Therefore, $\Sigma_c - \Sigma_b = \mathbb{E}[\text{Diag}(b) - bb^\top]$ is positive semidefinite. It follows that:

$$\Sigma_b \preccurlyeq \Sigma_c.$$

Here, equality holds if and only if model belief is almost surely deterministic for at least one alternative in $\mathcal{J}$.

It follows that the Loewner order of the covariance for the plug-in estimators obeys

$$D_g \Sigma_b D_g^\top \preccurlyeq D_g \Sigma_c D_g^\top$$

where equality holds if and only if model belief is almost surely deterministic for at least one alternative in $\mathcal{J}$.[8]

For $N$ independent runs,

$$\text{Var}\left(\hat{\theta}_b^{(N)}\right) = \frac{D_g \Sigma_b D_g^\top}{N} \quad \text{and} \quad \text{Var}\left(\hat{\theta}_c^{(N)}\right) = \frac{D_g \Sigma_c D_g^\top}{N},$$

which proves the sampling covariance result in the proposition.

For any target accuracy tolerance $\epsilon > 0$ and confidence level $1 - \delta$ with $0 < \delta < 1$, let $N_b$ be the minimal number of runs to guarantee $\Pr\left(\|\hat{\theta}_b^{(N)} - \bar{\theta}\| \leq \epsilon\right) \geq 1 - \delta$, and let $N_c$ be the minimal number of runs to guarantee $\Pr\left(\|\hat{\theta}_c^{(N)} - \bar{\theta}\| \leq \epsilon\right) \geq 1 - \delta$.

---

[8]We assume $D_g$ to be full row rank, otherwise the equality may hold even with $\Sigma_b \prec \Sigma_c$.

By Chebyshev's inequality,

$$N_b = \lceil \frac{\text{tr}(D_g \Sigma_b D_g^\top)}{\epsilon^2 \delta} \rceil \quad \text{and} \quad N_c = \lceil \frac{\text{tr}(D_g \Sigma_c D_g^\top)}{\epsilon^2 \delta} \rceil.$$

It follows that

$$N_b \leq N_c$$

where equality holds if and only if model belief is almost surely deterministic for at least one alternative in $\mathcal{J}$. This proves the convergence speed result in the proposition.

## *LLM Input Prompt: Template*

This section presents the input prompt template used in this LLM-based demand estimation study. The only variation across choice scenarios is the Pampers price, marked with [⋯] in the prompt template.

**System Prompt**

You are visiting a store to buy baby diapers. You see some diaper brands. Answer if you would buy any of these diaper brands and, if so, which brand you would buy. Do not choose multiple brands.

**User Prompt**

You see the following baby diaper brands in the store: Pampers and Huggies (in no particular order). Pampers diapers are generally described as soft and include a wetness indicator. Huggies diapers are often noted for having a snug fit and for helping to prevent leaks. The unit prices are [⋯] cents per Pampers diaper and 30 cents per Huggies diaper. You may choose one of these diaper brands to buy, or choose "neither" if you prefer other diaper brands. Question: Would you choose to buy Pampers, Huggies, or neither?

*LLM Output Extraction: Sample Code*

This section provides a sample Python code snippet demonstrating how to query the OpenAI Chat Completions API to obtain the outputs required for calculating model choice and model belief, respectively.

```python
# Python Code Sample for Calling OpenAI API
response = OpenAI.chat.completions.create(
    model='gpt-4o-2024-11-20',
    messages=messages,
    logprobs=True,
    top_logprobs=20,
    max_completion_tokens=200,
    temperature=1.0,
)
# Extract the model's textual output (for calculating model choice)
model_choice_text = response.choices[0].message.content
# Extract the log probability data (for calculating model belief)
logprob_data = response.choices[0].logprobs.content
```

The API call is configured with `logprobs=True` to retrieve the necessary token-level probabilities. We set the `top_logprobs` parameter to 20, the maximum value permitted by OpenAI's API at the time of this research. The goal is to capture the probabilities for as many alternatives as possible in the choice set $\mathcal{J}$ at the pivot token's position. Reassuringly, the token probability distribution is typically steep, dominated by a few high-probability tokens. Nevertheless, using the maximum value of 20 tokens ensures robust coverage and enables stable normalization over a consistent set of 20 tokens when computing model belief.

From the `response` object, model choice is determined by parsing the generated text found in `response.choices[0].message.content`. The data required to calculate model belief is extracted from the `logprobs.content` attribute. Model belief is determined by `logprobs.content` at the pivot token.

*Visualizing the Effect of Temperature on LLM Responses*

Figure 4 plots LLM responses in this demand estimation study, at the Pampers price of 31 cents, when the temperature varies from 0 to 1.8. When the temperature exceeds 1.8, the LLM begins producing erratic and nonsensical responses. This observation is consistent with the increased randomness in token sampling at high temperatures, which can propagate through the output sequence and result in incoherent or off-topic completions.

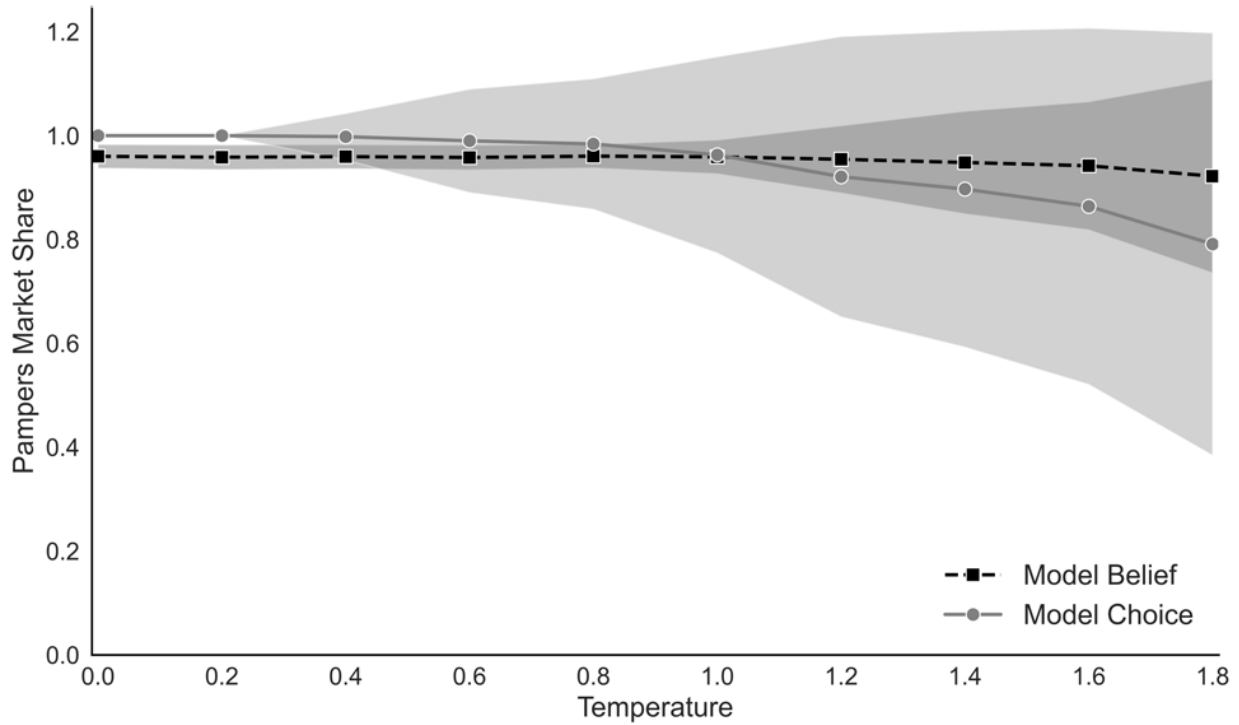

**Figure 4:** Effect of Temperature on Model Belief Versus Model Choice.

*Notes:* For each temperature value, the figure displays the model belief and model choice for Pampers at the price of 31 cents across 1,000 runs. The lines plot the mean values (Pampers market shares) and the shaded areas indicate one standard deviation from the mean, without capping the values at 1. At temperature = 1.0, model belief and model choice yield nearly identical mean values but model choice shows substantially greater standard deviation.